\documentclass{article} 
\usepackage{iclr2025_conference,times}

\usepackage{amsmath,amsfonts,bm}

\def\eqref#1{equation~\ref{#1}}

\def\1{\bm{1}}

\DeclareMathAlphabet{\mathsfit}{\encodingdefault}{\sfdefault}{m}{sl}
\SetMathAlphabet{\mathsfit}{bold}{\encodingdefault}{\sfdefault}{bx}{n}

\usepackage{newfloat}
\usepackage{listings}
\usepackage{newfloat}
\usepackage{listings}
\usepackage{moreverb}
\usepackage{mathtools}
\usepackage{amssymb}
\usepackage{graphics}
\usepackage{subfigure}
\usepackage{caption}
\usepackage{extarrows}
\usepackage{color}
\usepackage{framed}
\usepackage{bm}
\usepackage{mathrsfs}
\usepackage{mathabx}
\usepackage{multirow}
\usepackage{longtable}
\usepackage{paralist}
\usepackage{indentfirst}
\usepackage{relsize}
\usepackage{extarrows}
\usepackage{lineno,xcolor}
\usepackage{upgreek}
\usepackage{bm}
\usepackage{mwe}
\usepackage{xcolor}
\usepackage{booktabs}
\usepackage[flushleft]{threeparttable}
\usepackage{caption}
\usepackage{booktabs}
\usepackage{rotating} 
\usepackage{float}
\usepackage{nameref}

\usepackage{hyperref}
\usepackage{url}

\title{Dynamic Universal Approximation Theory: The Basic Theory for Residual-Based Deep Learning Computer Vision Models}

\author{Wei Wang \\
Department of Comp\\
The Hong Kong Polytechnic University\\
\texttt{weiuat.wang@connect.polyu.hk} \\
\And
Qing Li \\
Department of Comp \\
The Hong Kong Polytechnic University \\
\texttt{qing-prof.li@polyu.edu.hk} \\
}

\iclrfinalcopy 
\begin{document}

\maketitle

\begin{abstract}
Computer Vision (CV) is one of the most critical branches in artificial intelligence. In recent years, various deep learning models based on convolutional architectures and Transformers have been developed to address diverse challenges in CV. These algorithms have found practical applications in fields such as robotics and facial recognition. Despite the increasing capabilities of modern CV models, several fundamental questions remain unresolved. For instance, what underpins the generalization ability of Convolutional Neural Networks (CNNs)? The root of these issues lies in the lack of a robust theoretical foundation for deep learning models in CV. To address these key challenges, we propose the Dynamic Universal Approximation Theorem (DUAT), an advance of the Universal Approximation Theorem (UAT). We demonstrate that most residual-based deep learning models in CV fall under the category of DUAT functions. By doing so, we aim to clarify some theoretical issues in CV.
\end{abstract}

\section{Introduction}
\label{section:Introduction}
As a core branch of artificial intelligence, CV has a wide range of applications, encompassing tasks such as image segmentation~\cite{minaee2021image,wang2022medical}, classification~\cite{rawat2017deep,wang2017residual}, video synthesis~\cite{wang2018video,wang2019few}, and restoration~\cite{wang2019edvr,nah2019ntire}. This broad applicability highlights its enormous technical potential and practical significance, making the intelligent processing of CV tasks a highly important area of research. The most effective solutions in this field today are based on deep learning algorithms. 

One of the earliest and most influential works in using deep learning for CV was LeNet~\cite{726791}, which pioneered the use of CNNs for handwritten digit recognition, ushering in a new era of image processing. Following this, AlexNet's~\cite{10.1145/3065386} remarkable performance on the ImageNet~\cite{Russakovsky2014ImageNetLS} dataset not only achieved a significant leap in image classification accuracy but also cemented CNNs as the dominant method in visual processing. The subsequent introduction of ResNet~\cite{He_Zhang_Ren_Sun_2015} further improved image recognition accuracy and established residual-based structure as a foundational architecture for future network designs, influencing nearly all subsequent deep learning models~\cite{Ren2015FasterRT,Xie2016AggregatedRT,Szegedy2014GoingDW,Gao2019Res2NetAN,Liu2021SwinTH}.

In recent years, the success of Transformers~\cite{Vaswani2017AttentionIA} in the field of natural language processing (NLP) has gradually permeated CV. Researchers have begun designing models for CV based on Transformers, such as the Vision Transformer (ViT)~\cite{dosoViTskiy2020image}, Swin Transformer~\cite{Liu2021SwinTH} and StructViT~\cite{Kim2024LearningCS}, which have demonstrated capabilities comparable to CNNs in image tasks (For convenience, we collectively refer to all Transformer-based models in the CV domain as ViTs.). Currently, research in deep learning for image processing predominantly revolves around CNNs~\cite{Huang2023AreLK,Wang2022CanCB}, Transformers~\cite{Shi2023TransNeXtRF,Wang2023RepViTRM}, or strategies that integrate both~\cite{Lv2023DETRsBY,Hou2024SalienceDE}, continuously driving technological advancements.

Despite the significant advances in deep learning-based computer vision, several fundamental questions remain unanswered. For instance, why do CNNs require such deep architectures? What underpins the generalization ability of CNNs? Why do residual networks outperform fully convolutional networks like VGG~\cite{Simonyan2014VeryDC}? What are the core differences between residual-based CNNs and Transformer-based models? We believe that these questions arise from the lack of a foundational theory for CNNs.

While some researchers have attempted to explain CNNs through various approaches—such as using visualization techniques to analyze the relationship between feature maps and original images~\cite{Wei2016SelectiveCD} or exploring CNNs from the frequency domain perspective~\cite{yin2019fourier,xu2019training}—these explanations often have limitations and are somewhat subjective. Similarly, studies on the interpretability of ViTs~\cite{park2022vision,bai2022improving} suggest that the multi-head attention (MHA) mechanism in Transformers tends to capture low-frequency information, in contrast to CNNs, which function as high-pass filters. These studies advocate for combining the strengths of both models or improving ViTs to better capture high-frequency details. However, these methods only provide intuitive explanations for specific problems and fail to offer a theoretical framework for addressing broader issues within CNNs and ViTs. Essentially, they describe or interpret phenomena occurring within CNNs or ViTs rather than providing a comprehensive theory for these architectures.

To address these challenges, we propose the DUAT, which applies to both residual-based CNNs and ViTs. DUAT is an advancement of the UAT, specifically developed to address the commonly used residual structures in deep learning models for computer vision. Although previous studies~\cite{Lin2022OnTU, Yang2024OnTR, Zhou2018UniversalityOD, Zhou2020TheoryOD} have attempted to use UAT to explain certain aspects of CNNs, they are often limited by specific conditions, lacking broad applicability or practical guidance. For example, these studies have not fundamentally clarified the differences between residual-based CNNs and VGG, nor have they extended to ViTs.

Our objective is to employ the Matrix-Vector Method introduced in DUAT2LLMs~\cite{wang2024universalapproximationtheorybasic} to demonstrate that both residual-based CNNs and ViTs fall within the class of DUAT functions, theoretically resolving the aforementioned issues and providing practical insights for network design~\cite{wang2024universalapproximationtheoryfoundations}. Our contributions are as follows:

\begin{itemize}
\item We further developed the UAT to DUAT, which could help explain some problems in the CV area.

\item We demonstrate that the residual-based multi-layer convolutional and Transformer networks commonly employed in the field of CV belong to the DUAT function. 

\item We explained why CNNs require deep networks, considering the characteristics of image data and the DUAT perspective. 

\item We proved why the widely used residual structure in CV is so powerful beyond the full convolutional networks, like VGG. 

\item We generally compared a residual-based CNN and a ViT from the DUAT perspective, illustrating the source of their powerful capabilities and explaining why both models appear to have similar performance. 

\end{itemize}

The structure of this paper is as follows:
In Section \ref{section:UAT for CV}, we introduce the UAT and the Matrix-Vector Method, as well as DUAT.
In Section \ref{section:Matrix-Vector Method and Deep Learning}, we use the Matrix-Vector Method to transform some commonly used operations in CV into matrix-vector forms, such as 2D convolution (\ref{Sectio:Matrix-Vector Method for Conv2D}), 3D convolution (\ref{Sectio:Matrix-Vector Method for Conv3D}), mean-pooling (\ref{section:Matrix-Vector Method for Mean-Pooling}), and then we give the DUAT format of residual-based CNNs (\ref{section:The DUAT Format of Residual-Based CNNs}) and ViTs (\ref{section:Transformer for CV}).
In Section \ref{section:Discussion}, we address some fundamental questions in CV: why CNN networks need to be deep (\ref{section:Why do CNNs must be deep?}), why residual-based CNNs are more powerful than VGG (\ref{section:What makes ResNet to be the Winner in CV?}) and the differences between residual-based CNNs and ViTs (\ref{section:The differece between ResNet and ViT}).

\section{From UAT to DUAT}\label{section:UAT for CV}

\begin{figure}[t!]
\centering
\includegraphics[width=0.45\textwidth]{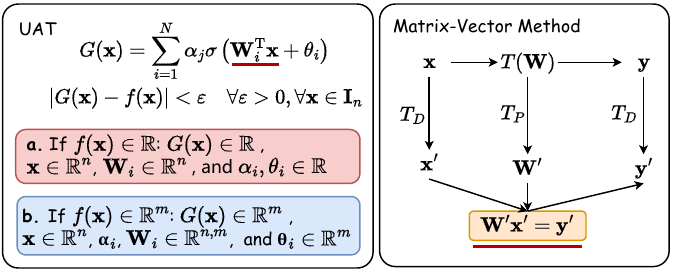}
\caption{The basic format of UAT and The Matrix-Vector Method and their relationship.}
\label{fig:UAT}
\end{figure}

Our goal is to unify residual-based CNNs and ViTs within the DUAT framework, where DUAT can be seen as an advanced form of UAT. Before introducing DUAT, we first provide an overview of UAT and briefly describe the Matrix-Vector Method proposed in DUAT2LLMs~\cite{wang2024universalapproximationtheorybasic}.

Figure \ref{fig:UAT} offers a simplified overview from DUAT2LLMs. Here, $G(\mathbf{x})$ represents the general form of UAT, which approximates Borel-measurable functions over closed intervals~\cite{Cybenko1989ApproximationBS, Hornik1989MultilayerFN}, with $\mathbf{I}_n$ denoting the n-dimensional unit cube. UAT extends from one-dimensional to multi-dimensional approximations, illustrated in Figures \ref{fig:UAT}.\textbf{a} and \ref{fig:UAT}.\textbf{b}. The fundamental computational unit of UAT is expressed as $\alpha_j\sigma(\mathbf{W}_j^{\mathrm{T}}\mathbf{x} + \theta_j)$. $\mathbf{W}_j^{\mathrm{T}}\mathbf{x} + \theta_j$ can be interpreted as convolution and pooling operations in CNNs, or as MHA and linear operations in ViTs, the specific mathematical forms of these basic operations differ from $\mathbf{W}_j^{\mathrm{T}}\mathbf{x} + \theta_j$. To bridge this gap, the matrix-vector approach is introduced.

The right of Figure \ref{fig:UAT} illustrates how these operations can be transformed into a matrix-vector multiplication form, $\mathbf{W}_j^{\mathrm{T}}\mathbf{x} + \theta_j$, using the Matrix-Vector Method. In this context, $T$ represents various transformations within CNNs and ViTs, while $T_D$ and $T_P$ denote specific matrix-vector transformations, varying with the choice of $T$. These transformations require converting input data $\mathbf{x}$, output data $\mathbf{y}$, and parameters $\mathbf{W}$ into their vectorized forms $\mathbf{x}'$, $\mathbf{y}'$, and parameter matrix $\mathbf{W}'$. These transformations ensure the computational relation $\mathbf{W}' \mathbf{x}' = \mathbf{y}'$, as shown in Figure \ref{fig:UAT}. We use a prime symbol $'$ to indicate the matrix-vector representations of these transformed variables, distinguishing them from their original forms. This establishes the link between basic network operations and $\mathbf{W}_j^{\mathrm{T}}\mathbf{x} + \theta_j$ (notably, we omit the bias term for simplification, focusing on basic operations' transformations, like convolution).

With this connection established between network basic operations and UAT's basic computational units, what about multi-layer networks? By expressing multi-layer networks mathematically in matrix-vector form, we observe that fully convolutional networks align with UAT in mathematical format (we illustrate this using the VGG network’s mathematical form in Section \ref{section:What makes ResNet to be the Winner in CV?}). As most current CV networks employ residual connections, writing these networks in corresponding mathematical terms reveals that their overall mathematical format aligns with UAT. However, unlike classic UAT, most parameters in these formulas vary with the input, making them functions of the input. We define this form of UAT, where some or all parameters depend on the input, as DUAT. (In Section \ref{section:The DUAT Format of Residual-Based CNNs} and \ref{section:Transformer for CV}, we present the DUAT mathematical form for a multi-layer residual-based CNN and ViT.) To demonstrate that a multi-layer network qualifies as DUAT, we use the matrix-vector method to represent the mathematical form of these networks, establish their consistency with UAT, and verify that certain UAT parameters vary dynamically with the input, confirming them as DUAT functions. Through the introduction of DUAT, we aim to address unresolved questions in computer vision.

\begin{figure}[t!]
\centering
\includegraphics[width=0.45\textwidth]{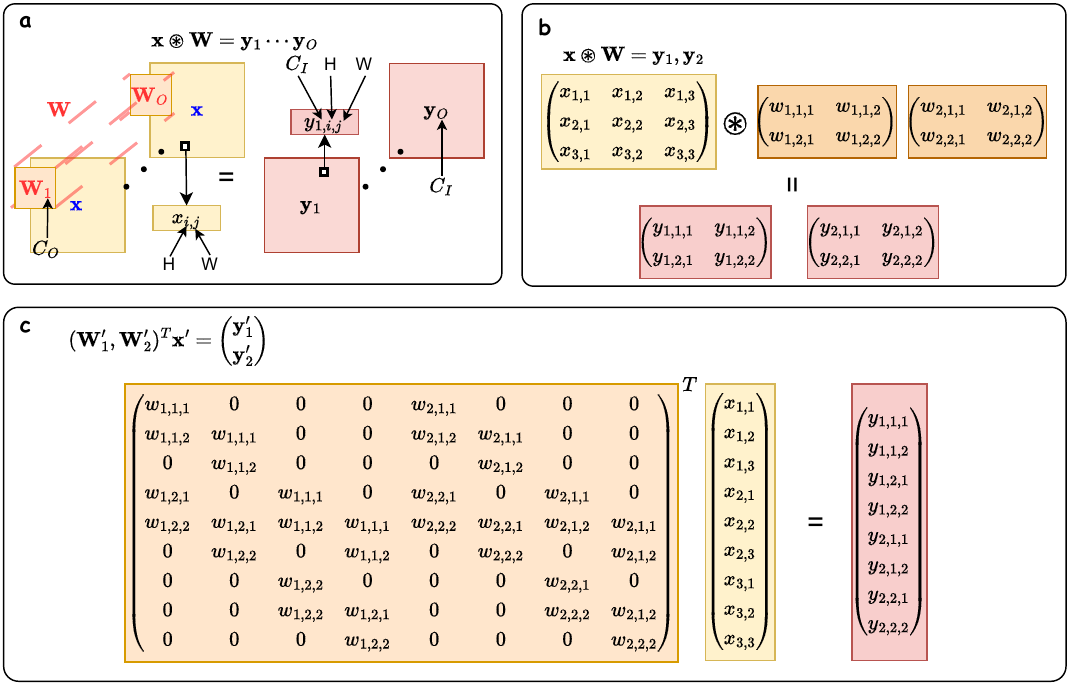}
\caption{
1-O Conv2D process and the matrix-vector transformation example. In \textbf{a}, we present the operation of 1-O Conv2D. At the top of \textbf{a}, \textbf{b} and \textbf{c}, we provide the mathematical formula corresponding to the diagram. Additionally, we group the variables in the figure, such as $\mathbf{W} = Concat(\mathbf{W}_1, \cdots, \mathbf{W}_O)$. In \textbf{b}, we provide an example corresponding to \textbf{a}, and in \textbf{c}, we show the matrix-vector form of the example given in \textbf{b}. The following figures follow the same conventions.}
\label{fig:Conv2D_1_O}
\end{figure}

\section{The DUAT Format of CNNs and ViTs}
\label{section:Matrix-Vector Method and Deep Learning}

In the previous sections, we explained that to unify CNNs and ViTs under the DUAT, it is necessary to demonstrate that their various transformations (convolution, mean pooling, MHA, and linear) can be represented in matrix-vector form. DUAT2LLMs has already shown how to represent MHA and linear in matrix-vector form. Therefore, in this section, we will specifically demonstrate how to convert convolution and mean pooling in CNNs into matrix-vector form using the Matrix-Vector Method. Additionally, we will interpret the unique aspects of Transformer within the context of CV. In the end, we give the  DUAT format of residual-based CNNs and ViTs.

Considering that convolution in CV can be either 2D or 3D and that the input and output may consist of single or multiple channels. We use the notation 1 to represent a single channel, I to represent an input with I channels, and O to represent an output with O channels. For example, 1-O indicates an operation with one channel input and $O$ channels output. Below, we will show how to convert these transformations into the corresponding matrix-vector forms. Then we will give the DUAT format of residual-based CNNs and ViTs.

\begin{figure}[t!]
\centering
\includegraphics[width=0.45\textwidth]{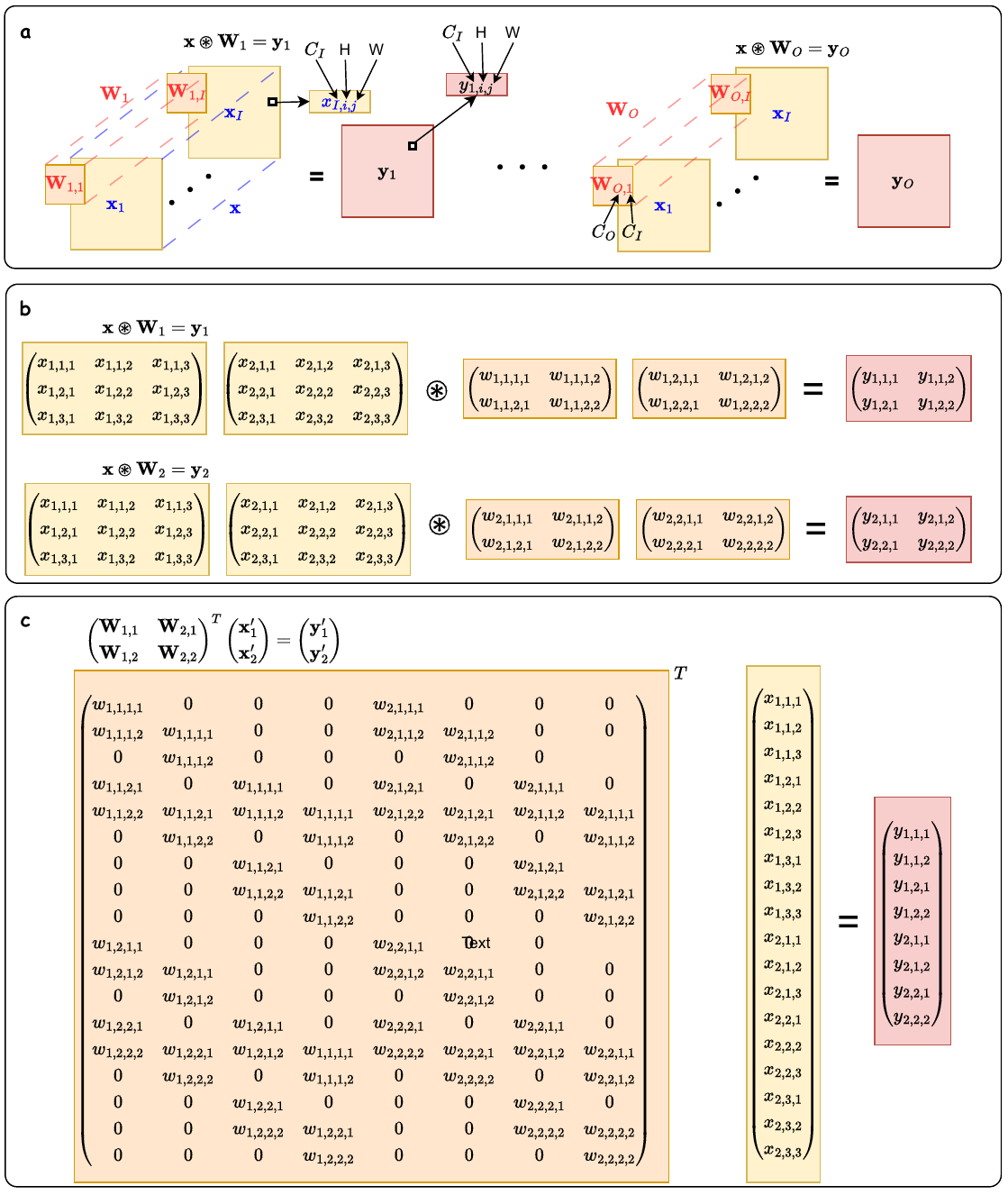}
\caption{I-O Conv2D process and the matrix-vector transformation example.}
\label{fig:Conv2D_I_O}
\vspace{-1.5em}
\end{figure}

\subsection{The Matrix-Vector Method for Conv2D}
\label{Sectio:Matrix-Vector Method for Conv2D}

In this section, we will detail how to convert 2D convolution into its corresponding matrix-vector forms, focusing on two scenarios: 1-O and I-O. To clearly describe the 2D convolution process, we first outline the general procedure and some basic conventions. The general process of 2D convolution involves convolving a kernel with the input data and then summing along the input channel direction (referred to as the $C_I$ direction) to obtain an element in the output. The convolution kernel slides along the height ($H$) and width ($W$) directions to produce the entire output feature map. To generate multiple feature maps, multiple sets of convolution kernels, $C_O$, are required, with each set having the same number of channels as the input channels. Depending on the context, $C_I$, $C_O$, $H$, and $W$ can represent either the corresponding dimension directions or the number of dimensions in those directions. Figures \ref{fig:Conv2D_1_O} and \ref{fig:Conv2D_I_O} illustrate the conversion process for the cases of 1-O and I-O, transforming them into matrix-vector representations.

Specifically, Figure \ref{fig:Conv2D_1_O}.\textbf{a} shows the process of 1-O Conv2D. Here, $\mathbf{W}_1 \cdots \mathbf{W}_O$ represent the convolution kernels, which slide across the input data along the $H$ and $W$ directions. For a single-channel input producing $O$ channels output, $O$ convolution kernels convolve with the same input $\mathbf{x}$ to produce outputs $\mathbf{y}_1 \cdots \mathbf{y}_O$. Figure \ref{fig:Conv2D_1_O}.\textbf{b} provides an example of a 1-O Conv2D. Additionally, Figure \ref{fig:Conv2D_1_O}.\textbf{c} further converts the example given in Figure \ref{fig:Conv2D_1_O}.\textbf{b} into its corresponding matrix-vector form. Therefore, it is straightforward to derive the matrix-vector form of 1-O Conv2D as follows:

\begin{equation}
\begin{aligned}
&\left(\begin{array}{cccc}
\mathbf{W}_1'&\mathbf{W}_2'&\cdots\mathbf{W}_O'
\end{array}\right)\diamond
\mathbf{x}'=\left(\begin{array}{cccc}
\mathbf{y}_1'\\
\mathbf{y}_2'\\
\vdots\\
\mathbf{y}_O'
\end{array}\right)
\end{aligned}
\label{eq:Conv2D_1_O}
\end{equation}

Figure \ref{fig:Conv2D_I_O}.\textbf{a} depicts the general process of I-O Conv2D, considering the scenario with $I$ channels input and $O$ channels output. This requires $O$ sets of convolution kernels, each set containing $I$ kernels, collectively represented as $\mathbf{W}_{1,1} \cdots \mathbf{W}_{1,I} \cdots \mathbf{W}_{O,1} \cdots \mathbf{W}_{O,I}$. Each kernel $\mathbf{W}_{i,j}$ with indices $i,j$ convolves with the $j$-th input channel $\mathbf{x}_{j}$ to produce an intermediate output $\mathbf{y}_{i,j}$. Summing all intermediate outputs $\mathbf{y}_{i,1} \cdots \mathbf{y}_{i,I}$ in the $i$-th set yields the final $i$-th output channel $\mathbf{y}_{i}$. Figure \ref{fig:Conv2D_I_O}.\textbf{b} provides a simplified example. Subsequently, Figure \ref{fig:Conv2D_I_O}.\textbf{c} presents the corresponding matrix-vector representation for the example in Figure \ref{fig:Conv2D_I_O}.\textbf{b}. Therefore, the matrix-vector formula for I-O Conv2D can be expressed as follows:

\begin{equation}
\begin{aligned}
&\left(\begin{array}{cccccc}
\mathbf{W}_{1,1}'&\mathbf{W}_{2,1}'&\cdots&\mathbf{W}_{O,1}'\\
\mathbf{W}_{1,2}'&\mathbf{W}_{2,2}'&\cdots&\mathbf{W}_{O,2}'\\
\vdots&\vdots&\vdots&\vdots\\
\mathbf{W}_{1,I}'&\mathbf{W}_{2,I}'&\cdots&\mathbf{W}_{O,I}'
\end{array}\right)\diamond
\left(\begin{array}{c}
\mathbf{x}_1'\\
\mathbf{x}_2'\\
\vdots\\
\mathbf{x}_I'
\end{array}\right)=\left(\begin{array}{c}
\mathbf{y}_1'\\
\mathbf{y}_2'\\
\vdots\\
\mathbf{y}_O'
\end{array}\right)
\end{aligned}
\label{eq:Conv2D_I_O}
\end{equation}

\begin{figure}[t!]
\centering
\includegraphics[width=0.45\textwidth]{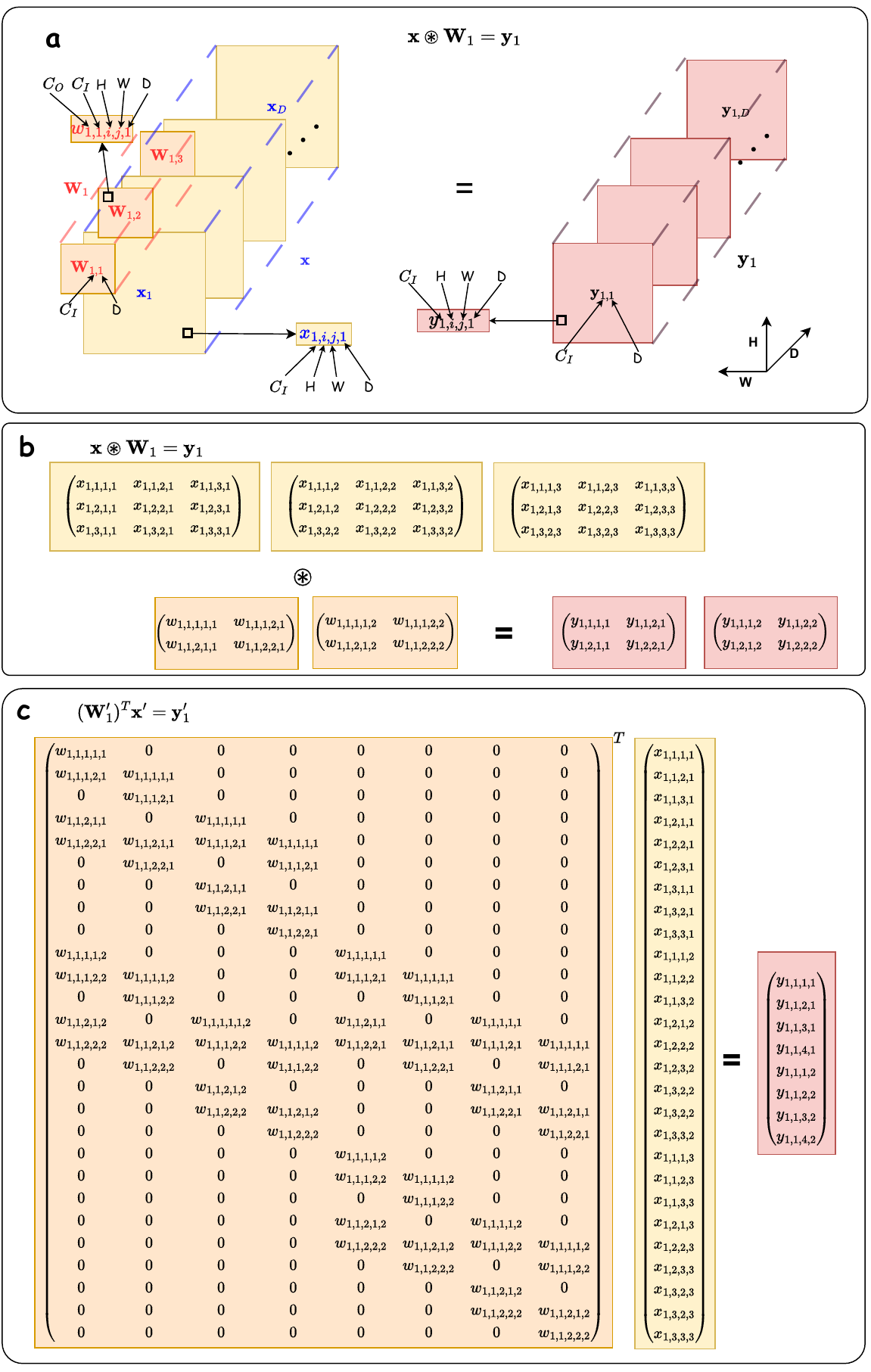}
\caption{1-O Conv3D process and the matrix-vector transformation example.}
\label{fig:Conv3D_1_O}
\vspace{-1.5em}
\end{figure}

\begin{figure*}[t!]
\centering
\includegraphics[width=1\textwidth]{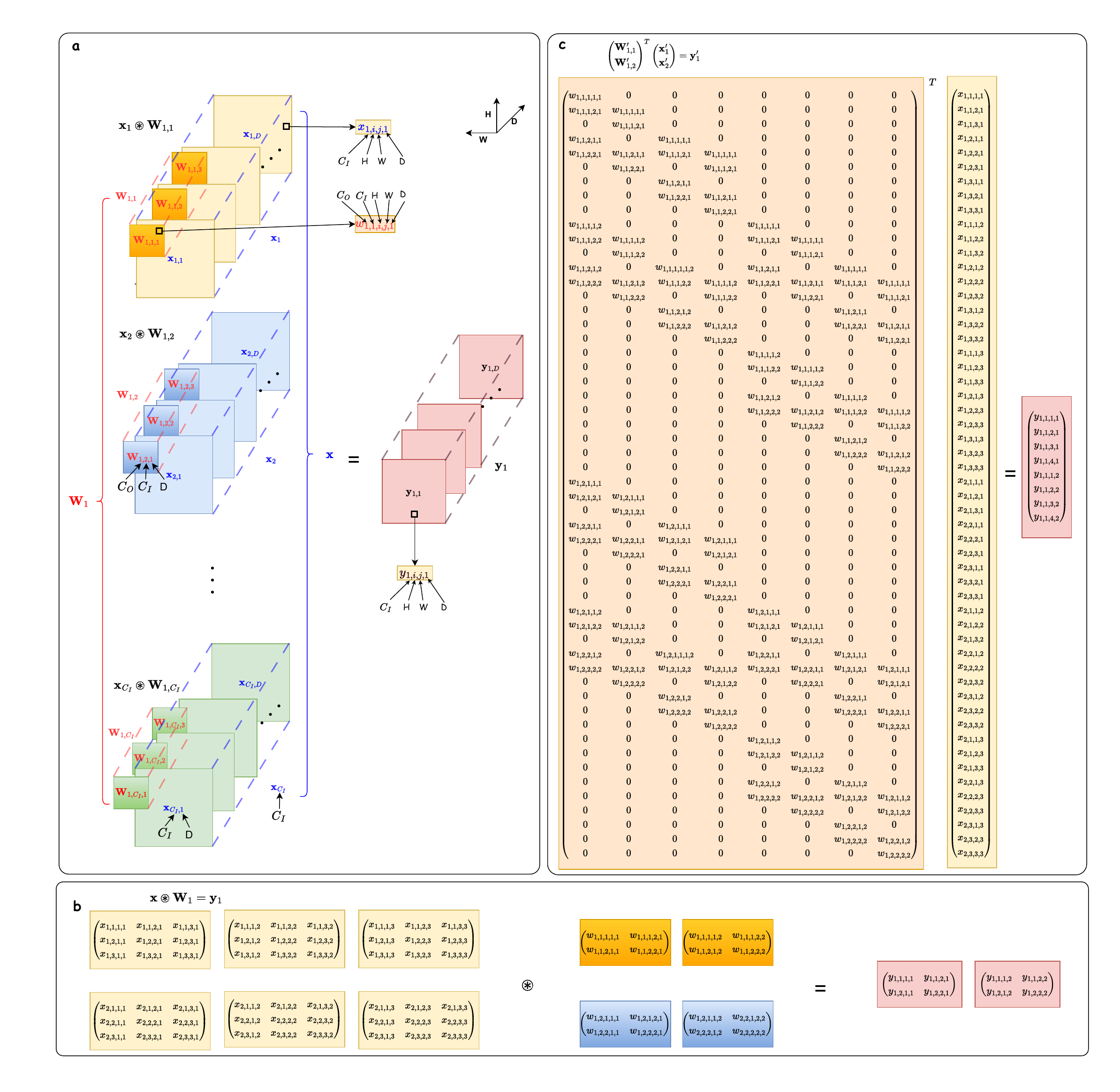}
\caption{I-1 Conv3D process and the matrix-vector transformation example.}
\label{fig:Conv3D_I_1}
\vspace{-1.5em}
\end{figure*}

\subsection{The Matrix-Vector Method for Conv3D}
\label{Sectio:Matrix-Vector Method for Conv3D}

3D convolution is a structure designed primarily for video or 3D image data, consisting mainly of two forms: 1-O Conv3D and I-O Conv3D. Given their complexity, we provide proof examples only for the 1-1 and I-1 Conv3D forms. The forms of 1-O and I-O Conv3D can be easily derived from the 1-1 and I-1 Conv3D forms. Compared with 2D convolution, 3D convolution has an additional depth dimension ($D$). So the convolution kernel could slide along with $H,W,D$, and it also needs to sum in $C_I$ direction.

Figure \ref{fig:Conv3D_1_O} illustrates the process of converting 1-1 Conv3D into its matrix-vector form. Figure \ref{fig:Conv3D_1_O}.\textbf{a} shows the general form of 1-O Conv3D. As a 3D convolution, the input data typically has four dimensions: $(C_{I}, H, W, D)$, representing the number of input channels, height, width, and depth, respectively. In 1-1 Conv3D, $C_{I}$ equal to 1. The convolution kernel has an additional dimension $C_{O}$, representing the number of output channels, and in this case, $C_{O} = 1$. The convolution kernel slides along the width (W), height (H), and depth (D) directions. Figure \ref{fig:Conv3D_1_O}.\textbf{b} presents a specific example, while Figure \ref{fig:Conv3D_1_O}.\textbf{c} converts this example into its matrix-vector form (ignoring changes in H, W, D dimensions due to lack of padding).

According to Figure \ref{fig:Conv3D_1_O}, the matrix-vector form of 1-O Conv3D can be easily derived and is consistent with Eq. \ref{eq:Conv2D_1_O}. The difference lies in the fact that $\mathbf{W}_{i}$ has depth, so the corresponding $\mathbf{W}_{i}'$ needs to be recombined as shown in Figure \ref{fig:Conv3D_1_O}.\textbf{c}.

Similarly, Figures \ref{fig:Conv3D_I_1}.\textbf{a}, \textbf{b}, and \textbf{c} illustrate the general process, a simple example, and the corresponding matrix-vector form of I-1 Conv3D, respectively. According to Figure \ref{fig:Conv3D_I_1}, the matrix-vector form of I-O Conv3D is consistent with 
Eq. \ref{eq:Conv2D_I_O}. The difference is that both $\mathbf{x}_{i}$ and $\mathbf{W}_{i}$ have depth, so the corresponding $\mathbf{x}_{i}'$ and $\mathbf{W}_{i}'$ need to be obtained as shown in Figure \ref{fig:Conv3D_I_1}.\textbf{c}.

\subsection{The Matrix-Vector Method for Mean Pooling}
\label{section:Matrix-Vector Method for Mean-Pooling}

\begin{figure}[t!]
\centering
\includegraphics[width=0.30\textwidth]{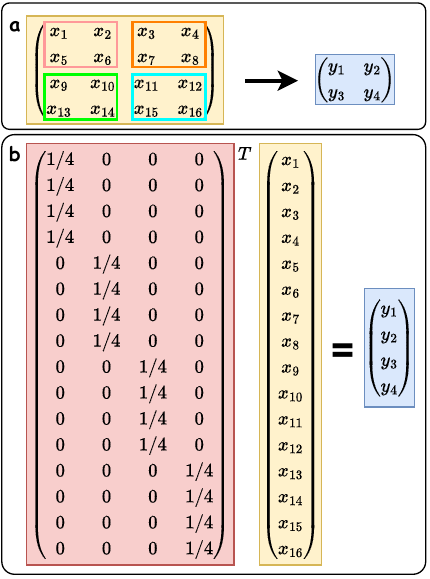}
\caption{Mean pooling process and the matrix-vector transformation example. Pink, orange, green and blue boxes represent mean pooling size.}
\label{fig:Mean-Pooling}
\end{figure}
Mean pooling is also an important technique in the field of CV, often used in conjunction with convolution. To demonstrate that CNNs can be expressed in the form of UAT, we need to prove that mean pooling can also be represented in matrix-vector form. In Figure \ref{fig:Mean-Pooling}, we provide an example. Figure \ref{fig:Mean-Pooling}.\textbf{a} shows the process of mean pooling, while Figure \ref{fig:Mean-Pooling}.\textbf{b} illustrates its corresponding matrix-vector form. Based on Figure \ref{fig:Mean-Pooling}, we can conclude that mean pooling can indeed be represented in matrix-vector form.

\subsection{The DUAT Format of Residual-Based CNNs}
\label{section:The DUAT Format of Residual-Based CNNs}
\begin{figure}[t!]
\centering
\includegraphics[width=0.35\textwidth]{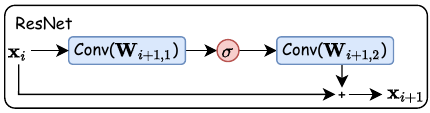}
\caption{A general residual module in ResNet.}
\label{fig:ResNet}
\vspace{-1.5em}
\end{figure}

Building on the previous derivation, we know that almost all types of basic operations in CNNs can be expressed in matrix-vector form. Here, we present the DUAT formulation for a general multi-layer residual-based CNN. Assuming a basic residual CNN module as shown in Figure \ref{fig:ResNet}, the DUAT form for a multi-layer residual-based CNN can be expressed as follows (for detailed deduction, refer to \textcolor{blue}{Appendix A.3}):

\begin{equation}
\begin{aligned}
\mathbf{x}_{i+1}'=(\mathbf{x}_0'+\mathbf{b}'_{i+1,2})+\Sigma_{j=1}^{i+1}\mathbf{W}'_{j,2}\sigma (\mathbf{W}'_{j,1}\mathbf{x}'_{0}+\mathbf{b}'_{j,1}) 
\end{aligned}
\label{eq:resnet}
\end{equation}

where $ i = 0, 1, 2, \ldots $. For $ i = 0 $, we have $ \mathbf{b}'_{1,1} = \mathbf{b}'_{1,1} $; and for $ i > 0 $, $ \mathbf{b}'_{j,1} $ is defined by $\mathbf{b}'_{j,1} = \left(\mathbf{W}'_{j,1} \mathbf{b}'_{j-1,2} + \mathbf{b}'_{j,1}\right) + \mathbf{W}'_{j,1} \sum_{k=1}^{j-1} \mathbf{W}'_{k,2} \sigma \left(\mathbf{W}'_{k,1} \mathbf{x}'_{0} + \mathbf{b}'_{k,1}\right).$ Thus, $ \mathbf{b}'_{j,1} $ can be seen as a $ j $-layer UAT and the input is $\mathbf{x}_0$. For convenience, we also consider the first term as part of the UAT.  Additionally, for the term $ \mathbf{b}'_{j+1,2} $ at layer $ i $, it is given by $\mathbf{b}'_{j+1,2} = \mathbf{b}'_{1,2} \cdots \mathbf{b}'_{j,2} + \mathbf{b}'_{j+1,2}, j=1,2\cdots i$. Eq. \ref{eq:resnet} aligns with the UAT form, as illustrated in Figure \ref{fig:UAT}. In this context, $ \mathbf{b}_{j,1} $ functions as a dynamic parameter that is approximated through UAT. Consequently, Eq. \ref{eq:resnet} satisfies the DUAT criterion, adopting a UAT form where the parameters dynamically adapt to the input.

\subsection{The DUAT Format of ViTs}
\label{section:Transformer for CV}

\begin{figure}[htbp!]
\centering
\includegraphics[width=0.4\textwidth]{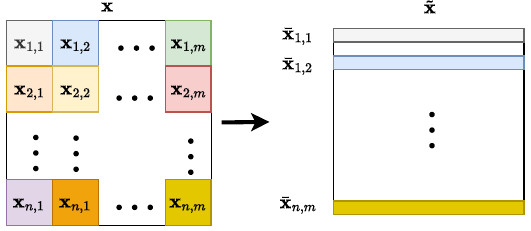}
\caption{The MHA operation in CV.}
\label{fig:MHA}
\end{figure}

In DUAT2LLMs, it has been demonstrated that the feedforward network (FFN) and MHA components in the Transformer architecture can be represented in matrix-vector form. Given that ViTs are also based on the Transformer framework, we only need to focus on the differences between Transformers in CV and those in LLMs.  A key feature of ViTs is that they divide the original image into multiple patches: $\mathbf{x}_{1,1}, \cdots, \mathbf{x}_{n,m}$. Each image patch is then reshaped into row vectors $\bar{\mathbf{x}}_{1,1}, \cdots, \bar{\mathbf{x}}_{n,m}$, which are concatenated along the column direction to obtain $\tilde{\mathbf{x}}$, as shown in Figure \ref{fig:MHA}. MHA and FFN operations are then applied, so aside from the initial processing part, the subsequent transformation process completely follows the description in DUAT2LLMs. 

\begin{figure}[htbp!]
\centering
\includegraphics[width=0.45\textwidth]{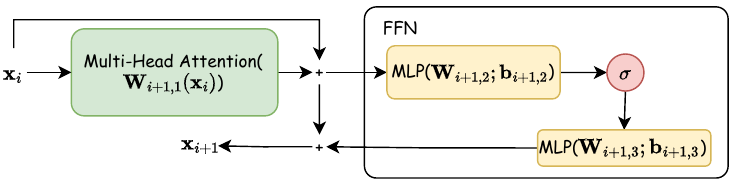}
\caption{The process of Transformer.}
\label{fig:TF}
\vspace{-0.5em}
\end{figure}

Therefore, a multi-layer ViT structured in Figure \ref{fig:TF} can be expressed as: 

\begin{equation}
\begin{aligned}
\mathbf{x}_{i+1}=(\mathbf{W}_{i+1,1}'\mathbf{x}_0+\mathbf{b}_{i+1,3}')+\sum_{j=1}^{i+1}\mathbf{W}'_{j,3}\sigma (\mathbf{W}'_{j,2}\mathbf{x}'_{0}+\mathbf{b}'_{j,2})
\end{aligned}
\label{eq:TF}
\end{equation}

When $ i = 0 $, the parameters are defined as follows: $ \mathbf{W}'_{1,1} = \mathbf{W}'_{1,1} $, $ \mathbf{b}'_{1,3} = \mathbf{b}'_{1,3} $, $ \mathbf{W}'_{1,3} = \mathbf{W}'_{1,3} $, $ \mathbf{W}'_{1,2} = \mathbf{W}'_{1,2} \mathbf{W}'_{1,1} $, and $ \mathbf{b}'_{1,2} = \mathbf{b}'_{1,2} $. We define the parameters for cases where $ i \geq 1 $. For each $ j = 1, 2, \ldots, i $, the updates are as follows: $ \mathbf{W}'_{j+1,1} = \mathbf{W}'_{j+1,1} \mathbf{W}_{j,1} $, $ \mathbf{b}'_{j+1,3} = \mathbf{W}'_{j+1,1} \mathbf{b}_{j,3} + \mathbf{b}'_{j+1,3} $, $ \mathbf{W}'_{j+1,2} = \mathbf{W}'_{j+1,2} \mathbf{W}_{j,1}$, and $ \mathbf{W}'_{j,3} = \mathbf{W}'_{j+1,1} \mathbf{W}'_{j,3}$. Additionally, for $ j = 2, \ldots, i + 1 $, the bias terms $ \mathbf{b}'_{j,2} $ are updated according to:
\begin{equation}
\begin{aligned}
\mathbf{b}'_{j,2} =& (\mathbf{W}'_{j,2} \mathbf{b}'_{j-1,3} + \mathbf{b}'_{j,2}) \\
+& \mathbf{W}'_{j,2} \sum_{k=1}^{j-1} \mathbf{W}'_{k,3} \sigma (\mathbf{W}'_{k,2} \mathbf{x}'_{0} + \mathbf{b}'_{k,2}).
\end{aligned}
\label{eq:b_j2}
\end{equation}

It means that the whole computing process is serial, if we want to compute $\mathbf{W}'_{3,1}$, we must calculate $\mathbf{W}'_{2,1}$, because $\mathbf{W}'_{3,1}=\mathbf{W}'_{3,1}\mathbf{W}'_{2,1}$. Thus, we have established that a multilayer ViT is also a DUAT function. The mathematical representation of the ViT aligns with the DUAT framework, with $\mathbf{b}'_{j,2}$ closely approximated by DUAT. The most notable difference between ViTs and residual-based CNNs is the dynamic input-dependent nature of the parameters in the MHA mechanism. Consequently, in the DUAT formulation associated with ViT, the parameters $\mathbf{W}'_{j,1}$ and $\mathbf{W}'_{j,2}$ for $i+1 \geq j \geq 1$, and $\mathbf{W}'_{j,3}$ for $i+1 > j > 1$ in layer $i$, all adapt dynamically based on the input. For further details, please refer to \textcolor{blue}{Appendix A.4}.

\section{Discussion}
\label{section:Discussion}
In this section, we will answer the questions in the field of CV proposed in the Introduction.

\subsection{Why must CNNs be Deep?}
\label{section:Why do CNNs must be deep?}
There are two main reasons why CNNs typically have deep layers. The first reason stems from the learning paradigm of CNNs. The design of CNNs is based on two well-known principles: local receptive fields (i.e., local features in images often contain important information) and spatial invariance (these important local features can appear anywhere in the image). These principles necessitate learning from each small region of the image, with all small regions sharing the same parameters. If parameters differed, it would imply that a certain object must appear in a specific part of the image, which is clearly unreasonable.

The above reasons dictate that image learning occurs in small patches, which explains why convolutional kernels are typically small. However, if the network has only a few layers, such as a single layer, it means that each small patch is learned independently. We know that objects in an image are generally composed of multiple such patches. Observing only a part of the image cannot provide a clear identification of the object. Therefore, a larger receptive field is needed, which can be achieved by increasing the number of layers to gather global information.

The second reason is derived from the DUAT. As shown in Figure \ref{fig:UAT}, increasing the number of layers in the network means a larger $N$, which brings the DUAT closer to approximating the target function. An image can be understood as a special function in a high-dimensional space, exhibiting strong correlations in 2D or 3D space. These two reasons are also applicable to ViTs. Figure \ref{fig:MHA} shows the preprocessing of ViTs, where $\tilde{\mathbf{x}}$ is the input to the ViTs. It is evident that each row shares the same parameters, and the learning of each row adheres to the principles of local receptive fields and spatial invariance.

\subsection{Why Residual-Based CNNs Excel in CV: Superior Generalization Ability}
\label{section:What makes ResNet to be the Winner in CV?}

In the previous discussion, we explained why deep networks are essential for learning in CNNs. However, why did VGG, despite being a deep network, not dominate the subsequent development of CV, whereas residual-based CNNs did? ResNet significantly shaped the design of later network architectures, with the residual structure appearing in many subsequent networks. What exactly endows the residual structure with such powerful capabilities? We think the answer is the generalization capability given by residual structure. To analyze the source of this power, we use the Matrix-Vector Method to express VGG and the residual-based CNNs into equations, and then compare them.

\begin{figure}[hbpt!]
\centering
\includegraphics[width=0.3\textwidth]{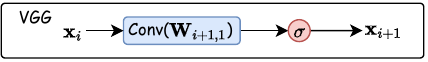}
\caption{VGG: The characters in '()' represent the corresponding operation's parameters.
}
\label{fig:VGG}
\end{figure}
Figure \ref{fig:VGG} illustrates a basic structure of VGG. Based on this figure, a $i$-layer VGG can be written as:

\begin{equation}
\begin{aligned}
\mathbf{x}'_{i} &=\sigma\{\mathbf{W}'_{i}\cdots\sigma\{\mathbf{W}'_{3}[\sigma(\mathbf{W}'_{2}\sigma(\mathbf{W}'_{1}\mathbf{x}'_{1}+\mathbf{b}'_{1})\\
&+\mathbf{b}'_{2})]+\mathbf{b}'_{3}\}+\mathbf{b}'_{i}\}
\end{aligned}
\label{eq:VGG}
\end{equation}

which perfectly aligns with the mathematical formulation of the multilayer UAT proposed by Hornik~\cite{Hornik1989MultilayerFN}. Therefore, based on Eqs. \ref{eq:VGG} and \ref{eq:resnet}, we can conclude that the fundamental reason ResNet is more powerful than VGG lies in the fact that the residual network is a DUAT function, while VGG is a UAT function. DUAT has a stronger dynamic approximation capability. As shown in Eq. \ref{eq:VGG}, once the VGG network finishes the training, its corresponding UAT parameters become fixed. This means that the VGG network can only approximate a single, fixed function. Although UAT has strong approximation capabilities, image data is highly variable, and the corresponding functions often change. 

In contrast, the residual network, corresponding to the DUAT in Eq. \ref{eq:resnet}, can dynamically adjust its bias parameters based on the input, allowing it to approximate the corresponding function dynamically. This adaptability is the fundamental source of the superiority of residual networks.

\subsection{The Difference between Residual-Based CNNs and ViTs}
\label{section:The differece between ResNet and ViT}

In this section, we will compare the DUAT format of Transformer-based ViTs and residual-based CNNs. Figure \ref{fig:TF} shows the general form of a Transformer. A multi-layer Transformer is shown in Eq \ref{eq:TF}. 
Comparing Eq. \ref{eq:TF} and \ref{eq:resnet}, the main difference between Transformer networks and residual-based multi-layer convolutional networks lies in how the input affects their corresponding DUAT parameters. In Transformer networks, input information influences both weight and bias in the corresponding DUAT. In contrast, in networks constructed with residual-based CNNs, input information only influences bias in the corresponding DUAT. Essentially, both types of networks have the ability to dynamically approximate functions based on input. This aligns with their nearly identical performance in the field of CV, supporting the theoretical foundation that both are DUAT functions and capable of dynamically approximating the corresponding functions based on input.

\section{Conclusion}
\label{section: Conclusion}
This paper delves into the theoretical foundations of deep learning in the field of CV. Specifically, the current CV landscape is primarily dominated by CNNs and Transformer models. We utilize the Matrix-Vector Method to unify these models under the framework of the DUAT. Based on this framework, we provide explanations for several common issues and techniques in CV. The requirement for deep networks in CNNs is jointly determined by the intrinsic characteristics of image data and the demands of DUAT. The robust generalization ability of residual networks stems from their ingenious design, which enables UAT to dynamically adapt to corresponding functions based on input data. Similarly, Transformer-based models possess this ability, with the key difference lying in which parameters of UAT they influence. Residual networks primarily affect the bias term, while Transformers influence both the weights and the bias. 

\section{Some Guesses on Human Vision Based on DUAT}

Based on the above discussion, we can consider that residual-based deep learning CV models are DUAT functions, which we refer to as DUAT vision. So, what is the relationship between human vision and DUAT vision? We believe that the human brain functions as a large network composed of one or multiple DUATs. Because first neural network is perceptron \cite{Rosenblatt1963PRINCIPLESON} which was designed based on the brain neuron. The Transformer and residual-based CNNs are DUAT functions, and the perceptron can be seen as a special form of UAT. Therefore, the concepts of memory and understanding, as defined by humans, can be categorized as reasoning based on the DUAT in our brains. Just as deep vision networks do not have a database storing images, they directly approximate the corresponding results from the inputs (which can be simplified as $y = DUAT(x)$). Specifically, humans and animals are born with certain weights in their brains, which are innate reflexes, such as infants instinctively knowing how to suckle. However, other weights in the brain need to be learned, and our process of observing the world from a young age is essentially a process of training these weights. Given that the eyes' vision operates at 30 frames per second \cite{lu2017high}, the brain would be trained on hundreds of millions of high-resolution images over a year. 

So, what is human memory? We believe it is the ability to generate corresponding results based on inputs and the weights learned by the brain. The so-called memory always has an anchor point that triggers it, i.e., something that evokes the memory. For example, an object might remind us of childhood memories because the visual input is similar or identical to what we saw in our childhood. We then derive a memory image, which is input back into the brain, and iterated repeatedly until the entire memory fragment is formed. Memory bias occurs when the brain's weight parameters are updated, leading to different results from the same input. Since the brain's weights are trained based on the natural world, it only infers results that are reasonable within the realm of cognition. This is why it is challenging for us to imagine things that do not exist, as the weights related to images in our brains are trained based on the natural world.

\bibliography{iclr2025_conference}
\bibliographystyle{iclr2025_conference}

\appendix
\clearpage
\setcounter{page}{1}
\setcounter{section}{0}

\section{The UAT Format of Residual-based CNNs and Transformer-based ViTs}

\textbf{Note: The sections \ref{section:The Properties of UAT}, Section \ref{section:From UAT to DUAT} and \ref{section:The UAT Format of Transformer-based ViTs} have already been proven in DUAT2LLMs. We also include them in this paper to ensure completeness.}

In this section, we will show the DUAT mathematical form corresponding to residual-based CNNs and Transformer-based ViTs (which also use residual connections by design). In Section \ref{section:The Properties of UAT}, we first introduce a lemma concerning UAT and then we introduce the DUAT in Section \ref{section:From UAT to DUAT}. Based on Section \ref{section:From UAT to DUAT}, we show that both residual-based CNNs (see Section \ref{section:The UAT Format of Residual-based CNNs}) and Transformer-based ViTs (see Section \ref{section:The UAT Format of Transformer-based ViTs}) are DUAT functions.

It is important to highlight that DUAT’s overall mathematical structure is aligned with UAT; the key distinction is that DUAT’s parameters are influenced by the input. Currently, this influence is achieved primarily through the use of residual connections. Therefore, we represent multi-layer residual networks in the corresponding UAT form and illustrate how their parameters are influenced by the input, establishing them as DUAT functions. Due to the close relationship between UAT and DUAT, some interchange of the terms UAT and DUAT may occur in the following sections.

\subsection{The Properties of UAT}
\label{section:The Properties of UAT}
Before expressing residual-based CNNs and Transformer-based ViTs in the UAT format, we present a lemma regarding UAT. There are two cases for UAT-approximated functions: $ f(\mathbf{x}) \in \mathbb{R} $ and $ f(\mathbf{x}) \in \mathbb{R}^m $. The proof for the case where $ f(\mathbf{x}) \in \mathbb{R} $ can be inferred from $ f(\mathbf{x}) \in \mathbb{R}^m $. Therefore, we will only provide the proof for approximating $ f(\mathbf{x}) \in \mathbb{R}^m $ using UAT.

Lemma 1. The mathematical form of UAT remains unchanged when multiplied by a matrix (constant).

\begin{equation}
\begin{aligned}
G(\mathbf{x}) &= \mathbf{\beta} \sum_{j  = 1}^N \mathbf{\alpha}_j \sigma\left(\mathbf{W}_j^{\mathrm{T}} \mathbf{x}+\mathbf{\theta}_j\right)\\
&= \sum_{j  = 1}^N \mathbf{\beta}\mathbf{\alpha}_j \sigma\left(\mathbf{W}_j^{\mathrm{T}} \mathbf{x}+\mathbf{\theta}_j\right)\\
\end{aligned}
\label{eq:UAT*W}
\end{equation}

Eq. \ref{eq:UAT*W} shows the representation of UAT multiplying a matrix. Let $ \mathbf{\alpha}_j = \mathbf{\beta}\mathbf{\alpha}_j $, and the general mathematical form of Eq. \ref{eq:UAT*W} remains consistent with the original UAT mathematical form. Thus, it is proven that the mathematical form of UAT remains unchanged when multiplied by a matrix (constant).

\subsection{From UAT to DUAT}
\label{section:From UAT to DUAT}
In this section, we first present a general form for a single-layer residual term in a network and demonstrate that the mathematical form of a multi-layer network composed of this residual term aligns with the DUAT framework. Before proceeding with the formal proof, we first define DUAT explicitly: DUAT shares the overall mathematical form of UAT, but with certain parameters influenced by the input, allowing them to change dynamically in response to it. We refer to these parameters as "dynamic parameters." In standard UAT, parameters remain fixed once training is complete, whereas, in DUAT, the dynamic parameters are functions of the input and thus vary with it. Consequently, dynamic parameters in DUAT may take the form of complex functions. Currently, DUAT primarily is implemented by residual structure, meaning that, in general, these complex functions are also DUAT. Next, we will give the proof.

A general residual term of the network can be written as: 

\begin{equation}
\begin{aligned}
\mathbf{x}'_{i} = (\mathbf{W}'_{i,1}\mathbf{x}'_{i-1} + \mathbf{b}'_{i,3}) + \mathbf{W}'_{i,3}\sigma (\mathbf{W}'_{i,2}\mathbf{x}'_{i-1} + \mathbf{b}'_{i,2})
\end{aligned}
\label{eq:res-term}
\end{equation}

where $i = 1, 2, 3, \ldots $ and $\mathbf{x}'_{0}$ represents the input. A multi-layer network structured in this way aligns with the mathematical form of DUAT. To demonstrate that a multi-layer network leverages the general term of Eq. \ref{eq:res-term} corresponds to DUAT, we first examine the forms of single-layer and two-layer networks, as shown in Eqs. \ref{eq:R-app-1} and \ref{eq:R-app-2}.

\begin{equation}
\centering
\begin{aligned}
\mathbf{x}'_{1} =(\mathbf{W}'_{1,1}\mathbf{x}'_{0}+\mathbf{b}'_{1,3})+\mathbf{W}'_{1,3}\sigma (\mathbf{W}'_{1,2}\mathbf{x}'_{0}+\mathbf{b}'_{1,2})
\end{aligned}
\label{eq:R-app-1}
\end{equation}

\begin{figure*}[htbp!]
\centering
\begin{equation}
\centering
\begin{aligned}
\mathbf{x}'_{2} & =\mathbf{W}'_{2,1}\mathbf{x}'_{1}+\mathbf{W}'_{2,3}\sigma (\mathbf{W}'_{2,2}\mathbf{x}'_{1}+\mathbf{b}'_{2,2})+\mathbf{b}'_{2,3}\\
&=(\mathbf{W}'_{2,1}\mathbf{x}'_{1}+\mathbf{b}'_{2,3})+\mathbf{W}'_{2,3}\sigma (\mathbf{W}'_{2,2}\mathbf{x}'_{1}+\mathbf{b}'_{2,2})\\
&=\{\mathbf{W}'_{2,1}[(\mathbf{W}'_{1,1}\mathbf{x}'_{0}+\mathbf{b}'_{1,3})+\mathbf{W}'_{1,3}\sigma (\mathbf{W}'_{1,2}\mathbf{x}'_{0}+\mathbf{b}'_{1,2})]+\mathbf{b}'_{2,3}\}\\
&+\mathbf{W}'_{2,3}\sigma \{\mathbf{W}'_{2,2}[(\mathbf{W}'_{1,1}\mathbf{x}'_{0}+\mathbf{b}'_{1,3})+\mathbf{W}'_{1,3}\sigma (\mathbf{W}'_{1,2}\mathbf{x}'_{0}+\mathbf{b}'_{1,2})]+\mathbf{b}'_{2,2}\}\\
&=\{\mathbf{W}'_{2,1}(\mathbf{W}'_{1,1}\mathbf{x}'_{0}+\mathbf{b}'_{1,3})+\mathbf{b}'_{2,3}+\mathbf{W}'_{2,1}\mathbf{W}'_{1,3}\sigma (\mathbf{W}'_{1,2}\mathbf{x}'_{0}+\mathbf{b}'_{1,2})\}\\
&+\mathbf{W}'_{2,3}\sigma \{\mathbf{W}'_{2,2}({\mathbf{W}}'_{1,1}\mathbf{x}'_{0}+\mathbf{b}'_{1,3})+\mathbf{b}'_{2,2}+\mathbf{W}'_{2,2}\mathbf{W}'_{1,3}\sigma (\mathbf{W}'_{1,2}\mathbf{x}'_{0}+\mathbf{b}'_{1,2})\}\\
&=\{(\underline{\mathbf{W}'_{2,1}\mathbf{W}'_{1,1}}\mathbf{x}'_{0}+\underline{\mathbf{W}'_{2,1}\mathbf{b}'_{1,3}+\mathbf{b}'_{2,3}})+\underline{\mathbf{W}'_{2,1}\mathbf{W}'_{1,3}}\sigma (\mathbf{W}'_{1,2}\mathbf{x}'_{0}+\mathbf{b}'_{1,2})\}\\
&+\mathbf{W}'_{2,3}\sigma \{\underline{\mathbf{W}'_{2,2}\mathbf{W}'_{1,1}}\mathbf{x}'_{0}+\underline{(\mathbf{W}'_{2,2}\mathbf{b}'_{1,3}+\mathbf{b}'_{2,2})+\mathbf{W}'_{2,2}\mathbf{W}'_{1,3}\sigma (\mathbf{W}'_{1,2}\mathbf{x}'_{0}+\mathbf{b}'_{1,2})}\}\\
\end{aligned}
\label{eq:R-app-2}
\end{equation}
\end{figure*}

In Eq. \ref{eq:R-app-2}, let $ \mathbf{W}'_{2,1}=\mathbf{W}'_{2,1}\mathbf{W}'_{1,1} $,
$ \mathbf{b}'_{2,1}=\mathbf{W}'_{2,1}\mathbf{b}'_{1,3}+\mathbf{b}'_{2,3} $,
$ \mathbf{W}'_{1,3}=\mathbf{W}'_{2,1}\mathbf{W}'_{1,3} $,
$ \mathbf{W}'_{2,2}=\mathbf{W}'_{2,2}\mathbf{W}'_{1,1} $, and
$ \mathbf{b}'_{2,2}=(\mathbf{W}'_{2,2}\mathbf{b}'_{1,3}+\mathbf{b}'_{2,2})+\mathbf{W}'_{2,2}\mathbf{W}'_{1,3}\sigma (\mathbf{W}'_{1,2}\mathbf{x}'_{0}+\mathbf{b}'_{1,2}) $. Thus, Eq. \ref{eq:R-app-2} can be written into Eq. \ref{eq:UAT-2-layer}.

\begin{equation}
\begin{aligned}
\mathbf{x}'_{2}&=(\mathbf{W}'_{2,1}\mathbf{x}'_{0}+\mathbf{b}'_{2,1})+\mathbf{W}'_{1,3}\sigma (\mathbf{W}'_{1,2}\mathbf{x}'_{0}+\mathbf{b}'_{1,2})\\
&+\mathbf{W}'_{2,3}\sigma (\mathbf{W}'_{2,2}\mathbf{x}'_{0}+\mathbf{b}'_{2,2})
\end{aligned}
\label{eq:UAT-2-layer}
\end{equation}

For the cleatity, we could write Eq. \ref{eq:R-app-1} and Eq. \ref{eq:UAT-2-layer} into Eq. \ref{eq:UAT-1-layer_R} and Eq. \ref{eq:UAT-2-layer_R}, where $UAT^R_1=\mathbf{W}'_{1,3}\sigma (\mathbf{W}'_{1,2}\mathbf{x}'_{0}+\mathbf{b}'_{1,2})$ and $UAT^R_2=\Sigma_{j=1}^2\mathbf{W}'_{j,3}\sigma (\mathbf{W}'_{j,2}\mathbf{x}'_{0}+\mathbf{b}'_{j,2})
$.

\begin{equation}
\begin{aligned}
\mathbf{x}'_{1} =(\mathbf{W}'_{1,1}\mathbf{x}'_{0}+\mathbf{b}'_{1,3})+UAT^R_1
\end{aligned}
\label{eq:UAT-1-layer_R}
\end{equation}

\begin{equation}
\begin{aligned}
\mathbf{x}'_{2}&=(\mathbf{W}'_{2,1}\mathbf{x}'_{0}+\mathbf{b}'_{2,1})+UAT^R_2
\end{aligned}
\label{eq:UAT-2-layer_R}
\end{equation}

According to Eq. \ref{eq:UAT-1-layer_R} and Eq. \ref{eq:UAT-2-layer_R}, aside from the initial term, the overall mathematical forms of the residual terms $ UAT^R_1 $ and $ UAT^R_2 $ are consistent with UAT. However, for $ UAT^R_2 $, its parameter $ \mathbf{b}'_{2,2} $ is influenced by the input, while other weight parameters, such as $ \mathbf{W}'_{2,1} $, can be disregarded. These parameters are only affected by other parameters, so once the network training is complete, they are essentially fixed. This allows us to focus primarily on the dynamic parameters. In Eq. $\mathbf{b}'_{2,2} = (\mathbf{W}'_{2,2}\mathbf{b}'_{1,3} + \mathbf{b}'_{2,2}) + \mathbf{W}'_{2,2}\mathbf{W}'_{1,3} \sigma (\mathbf{W}'_{1,2}\mathbf{x}'_{0} + \mathbf{b}'_{1,2}) $, except for the initial term, $ \mathbf{b}'_{2,2} $ also follows the same mathematical form as UAT. This can be interpreted as dynamically adjusting the bias term $ \mathbf{b}'_{2,2} $ via UAT based on the input. In conclusion, we have demonstrated that the mathematical forms of one-layer and two-layer residual networks are consistent with the UAT framework and the two-layer residual network is the DUAT function.

Next, we will use mathematical induction to prove that the mathematical form of a multi-layer residual network is also a DUAT function. Assume that the overall mathematical form of the first $ i $ layers of the residual network aligns with UAT. Our goal is to show that the overall mathematical form of the $ i+1 $-th layer remains consistent with UAT.

For clarity, we make the following definitions: since the overall mathematical form of the first $ i $ layers is consistent with UAT, we can write it as $ \mathbf{x}'_{i} = (\mathbf{W}'_{i,1} \mathbf{x}'_{0} + \mathbf{b}'_{i,1}) + UAT^R_i $, where the first term is explicitly written, and the remainder is denoted as $ UAT^R_i $, with $ UAT_{i}^R = \sum_{j=1}^{i} \mathbf{W}'_{j,2} \sigma (\mathbf{W}'_{j,1} \mathbf{x}'_{0} + \mathbf{b}'_{j,1}) $. Since $ \mathbf{x}'_{i+1} = (\mathbf{W}'_{i+1,1} \mathbf{x}'_{i} + \mathbf{b}'_{i+1,3}) + \mathbf{W}'_{i+1,3} \sigma (\mathbf{W}'_{i+1,2} \mathbf{x}'_{i} + \mathbf{b}'_{i+1,2}) $, we divide $ \mathbf{x}'_{i+1} $ into two parts: $ (\mathbf{W}'_{i+1,1} \mathbf{x}'_{i} + \mathbf{b}'_{i+1,3}) $ and $ \mathbf{W}'_{i+1,3} \sigma (\mathbf{W}'_{i+1,2} \mathbf{x}'_{i} + \mathbf{b}'_{i+1,2}) $.

First, consider $ (\mathbf{W}'_{i+1,1} \mathbf{x}'_{i} + \mathbf{b}'_{i+1,3}) $. Substituting $ \mathbf{x}'_{i} = (\mathbf{W}'_{i,1} \mathbf{x}'_{0} + \mathbf{b}'_{i,1}) + UAT^R_i $, we obtain Eq. \ref{eq:i+1_term_1}. Setting $ \mathbf{W}'_{i+1,1} = \mathbf{W}'_{i+1,1} \mathbf{W}'_{i,1} $ and $ \mathbf{b}'_{i+1,1} = \mathbf{W}'_{i+1,1} \mathbf{b}'_{i,1} + \mathbf{b}'_{i+1,3} $, we can simplify the first part to $ (\mathbf{W}'_{i+1,1} \mathbf{x}'_{0} + \mathbf{b}'_{i+1,1}) + \mathbf{W}'_{i+1,1} UAT^R_i $. Since the overall mathematical form of $ UAT^R_i $ is consistent with the $ i $-layer UAT and we have shown that the UAT form is preserved when multiplied by a matrix, we have thus demonstrated that the overall mathematical form of the first part remains consistent with UAT.

\begin{figure*}[htbp!]
\centering
\begin{equation}
\begin{aligned}
&(\mathbf{W}'_{i+1,1}\mathbf{x}'_{i} + \mathbf{b}'_{i+1,3})\\
=&\{\mathbf{W}'_{i+1,1}[(\mathbf{W}'_{i,1}\mathbf{x}'_{0}+\mathbf{b}'_{i,1})+UAT^R_i] + \mathbf{b}'_{i+1,3}\}\\
=&\{\mathbf{W}'_{i+1,1}(\mathbf{W}'_{i,1}\mathbf{x}'_{0}+\mathbf{b}'_{i,1})+\mathbf{W}'_{i+1,1}UAT^R_i + \mathbf{b}'_{i+1,3}\}\\
=&\{[\underline{ \mathbf{W}'_{i+1,1}\mathbf{W}'_{i,1}}\mathbf{x}'_{0}+\underline{(\mathbf{W}'_{i+1,1}\mathbf{b}'_{i,1}+ \mathbf{b}'_{i+1,3})}]+\mathbf{W}'_{i+1,1}UAT^R_i \}\\
\end{aligned}
\label{eq:i+1_term_1}
\end{equation}
\end{figure*}

Next, we show that the overall mathematical form of the second part, $ \mathbf{W}'_{i+1,3} \sigma (\mathbf{W}'_{i+1,2} \mathbf{x}'_{i} + \mathbf{b}'_{i+1,2}) $, is also consistent with UAT. Substituting $ \mathbf{x}'_{i} = (\mathbf{W}'_{i,1} \mathbf{x}'_{0} + \mathbf{b}'_{i,1}) + UAT^R_i $ into this expression, we arrive at Eq. \ref{eq:i+1_term_2}. Setting $ \mathbf{W}'_{i+1,2} = \mathbf{W}'_{i+1,2} \mathbf{W}'_{i,1} $ and $ \mathbf{b}'_{i+1,2} = (\mathbf{W}'_{i+1,2} \mathbf{b}'_{i,1} + \mathbf{b}'_{i+1,2}) + \mathbf{W}'_{i+1,2} UAT^R_i $, we can rewrite the second part as $ \mathbf{W}'_{i+1,3} \sigma (\mathbf{W}'_{i+1,2} \mathbf{x}'_{0} + \mathbf{b}'_{i+1,2}) $, which can be interpreted as a term in UAT. Here, $ \mathbf{b}'_{i+1,2} $ serves as a bias term approximated using DUAT.

\begin{figure*}[htbp!]
\centering
\begin{equation}
\begin{aligned}
&\mathbf{W}'_{i+1,3}\sigma (\mathbf{W}'_{i+1,2}\mathbf{x}'_{i} + \mathbf{b}'_{i+1,2})\\
=&\mathbf{W}'_{i+1,3}\sigma \{\mathbf{W}'_{i+1,2}[(\mathbf{W}'_{i,1}\mathbf{x}'_{0}+\mathbf{b}'_{i,1})+UAT^R_i] + \mathbf{b}'_{i+1,2}\}\\
=&\mathbf{W}'_{i+1,3}\sigma \{\mathbf{W}'_{i+1,2}(\mathbf{W}'_{i,1}\mathbf{x}'_{0}+\mathbf{b}'_{i,1})
+\mathbf{W}'_{i+1,2}UAT^R_i + \mathbf{b}'_{i+1,2}\}\\
=&\mathbf{W}'_{i+1,3}\sigma \{\underline{\mathbf{W}'_{i+1,2}\mathbf{W}'_{i,1}}\mathbf{x}'_{0}+\underline{(\mathbf{W}'_{i+1,2}\mathbf{b}'_{i,1}+ \mathbf{b}'_{i+1,2})+\mathbf{W}'_{i+1,2}UAT^R_i} \}\\
\end{aligned}
\label{eq:i+1_term_2}
\end{equation}
\end{figure*}

Let $ UAT^R_{i+1} = \mathbf{W}'_{i+1,1} UAT^R_i + \mathbf{W}'_{i+1,3} \sigma (\mathbf{W}'_{i+1,2} \mathbf{x}'_{0} + \mathbf{b}'_{i+1,2}) $. Then, we can express $\mathbf{x}'_{i+1}$ as $ \mathbf{x}'_{i+1} = (\mathbf{W}'_{i+1,1} \mathbf{x}'_{0} + \mathbf{b}'_{i+1,1}) + UAT^R_{i+1}$. Thus, apart from the initial term, the overall mathematical form of $\mathbf{x}'_{i+1}$ remains consistent with UAT. Since a single term does not alter the nature of UAT, we conclude that the overall form of a multi-layer residual network aligns with UAT. Furthermore, as certain bias parameters within the network, such as $\mathbf{b}'_{i+1,2}$, are also approximated using UAT, we have therefore demonstrated that the mathematical form of a multi-layer residual network is a UAT function.

\subsection{The DUAT Format of Residual-based CNNs}
\label{section:The UAT Format of Residual-based CNNs}

To demonstrate that the residual-based CNN is indeed a DUAT function, we present its general form as defined by Figure 8 in the main text, represented by Eq. \ref{eq:UAT-res-cnn}. Comparing this with the general form provided in Section \ref{section:From UAT to DUAT}, specifically $ \mathbf{x}'_{i+1} = (\mathbf{W}'_{i+1,1} \mathbf{x}'_{i} + \mathbf{b}'_{i+1,3}) + \mathbf{W}'_{i+1,3} \sigma (\mathbf{W}'_{i+1,2} \mathbf{x}'_{i} + \mathbf{b}'_{i+1,2}) $, we find their only distinction lies in the term $ \mathbf{W}'_{i+1,1} \mathbf{x}'_{i} + \mathbf{b}'_{i+1,3} $. It is straightforward to deduce that the presence or absence of $ \mathbf{W}'_{i+1,1} $ does not affect the overall mathematical form. Therefore, the residual-based CNN also qualifies as a DUAT function.

\begin{equation}
\begin{aligned}
\mathbf{x}'_{i+1} & = \mathbf{x}_i'+\mathbf{W}_{i+1,2}'\sigma(\mathbf{W}_{i+1,1}'\mathbf{x}'_i+\mathbf{b}_{i+1,1})+\mathbf{b}_{i+1,2}'\\
& = (\mathbf{x}_i'+\mathbf{b}_{i+1,2}')+\mathbf{W}_{i+1,2}'\sigma(\mathbf{W}_{i+1,1}'\mathbf{x}'_i+\mathbf{b}_{i+1,1}')\\
\end{aligned}
\label{eq:UAT-res-cnn}
\end{equation}

To clearly present the DUAT form of residual-based CNNs, we apply the method described in Section \ref{section:From UAT to DUAT}. Assuming that the general mathematical form of $\mathbf{x}'_i$ aligns with UAT, we decompose $\mathbf{x}'_i$ into a primary term and a remainder term, expressed as $\mathbf{x}'_i = (\mathbf{x}_0' + \mathbf{b}_{i,2}') + UAT^R_i$, where $ UAT_{i}^R = \sum_{j=1}^{i} \mathbf{W}'_{j,2} \sigma (\mathbf{W}'_{j,1} \mathbf{x}'_{0} + \mathbf{b}'_{j,1}) $ for $ i \geq 1 $. Substituting this into $ \mathbf{x}'_{i+1} = (\mathbf{x}_i' + \mathbf{b}_{i+1,2}') + \mathbf{W}_{i+1,2}' \sigma(\mathbf{W}_{i+1,1}' \mathbf{x}'_i + \mathbf{b}_{i+1,1}') $, we derive Eq. \ref{eq:UAT-res-cnn-2}. 

By setting $ \mathbf{b}_{i+1,2}' = \mathbf{b}_{i,2}' + \mathbf{b}_{i+1,2}' $ and $ \mathbf{b}_{i+1,1}' = (\mathbf{W}_{i+1,1}' \mathbf{b}_{i,1}' + \mathbf{b}_{i+1,1}') + \mathbf{W}_{i+1,1}' UAT^R_i $, we obtain $ \mathbf{x}_{i+1}' = (\mathbf{x}_0' + \mathbf{b}_{i+1,2}') + UAT^R_i + \mathbf{W}_{i+1,2}' \sigma(\mathbf{W}_{i+1,1}' \mathbf{x}_0' + \mathbf{b}_{i+1,1}') $. Given that the overall mathematical format of $ UAT^R_i $ is consistent with the UAT formula, it follows that $ \mathbf{x}_{i+1}' $ also conforms to the UAT format. 

Furthermore, we define $ UAT^R_{i+1} = UAT^R_i + \mathbf{W}_{i+1,2}' \sigma(\mathbf{W}_{i+1,1}' \mathbf{x}_0' + \mathbf{b}_{i+1,1}') $, leading to $ \mathbf{x}_{i+1}' = (\mathbf{x}_0' + \mathbf{b}_{i+1,2}') + UAT^R_{i+1} $, where $ UAT_{i+1}^R = \sum_{j=1}^{i+1} \mathbf{W}'_{j,2} \sigma (\mathbf{W}'_{j,1} \mathbf{x}'_{0} + \mathbf{b}'_{j,1}) $. Thus, $i+1$-layer residual-based CNN is:

\begin{equation}
\begin{aligned}
\mathbf{x}_{i+1}'=(\mathbf{x}_0'+\mathbf{b}'_{i+1,2})+\Sigma_{j=1}^{i+1}\mathbf{W}'_{j,2}\sigma (\mathbf{W}'_{j,1}\mathbf{x}'_{0}+\mathbf{b}'_{j,1}) 
\end{aligned}
\label{eq:resnet-app}
\end{equation}

where $ i = 0, 1, 2, \ldots $. For $ i = 0 $, we have $ \mathbf{b}'_{1,1} = \mathbf{b}'_{1,1} $; and for $ i > 0 $, $ \mathbf{b}'_{j,1} $ is defined by $\mathbf{b}'_{j,1} = \left(\mathbf{W}'_{j,1} \mathbf{b}'_{j-1,2} + \mathbf{b}'_{j,1}\right) + \mathbf{W}'_{j,1} \sum_{k=1}^{j-1} \mathbf{W}'_{k,2} \sigma \left(\mathbf{W}'_{k,1} \mathbf{x}'_{0} + \mathbf{b}'_{k,1}\right).$ Thus, $ \mathbf{b}'_{j,1} $ can be seen as a $ j $-layer UAT and the input is $\mathbf{x}_0$. For convenience, we also consider the first term as part of the UAT.  Additionally, for the term $ \mathbf{b}'_{j+1,2} $ at layer $ i $, it is given by $\mathbf{b}'_{j+1,2} = \mathbf{b}'_{1,2} \cdots \mathbf{b}'_{j,2} + \mathbf{b}'_{j+1,2}, j=1,2\cdots i$. Naturally, in a multilayer network, $\mathbf{b}'_{i,2},i\geq1$ within the corresponding UAT will also change. However, as shown in the derivation above, these adjustments involve only matrix addition of predefined parameters. Therefore, they can essentially be understood as fixed parameters and do not fundamentally alter the mathematical form of the UAT. Consequently, we will not elaborate on the specific forms of these parameters here. Some examples are provided in Figure \ref{fig:Conv_UAT}. Since $\mathbf{b}'_{j,1}$ adapts dynamically to the input, the multilayer convolutional network based on residuals can essentially be understood as a DUAT function.

\begin{figure*}[htbp!]
\centering
\begin{equation}
\begin{aligned}
&(\mathbf{x}_i'+\mathbf{b}_{i+1,2})+\mathbf{W}_{i+1,2}'\sigma(\mathbf{W}_{i+1,1}'\mathbf{x}'_i+\mathbf{b}_{i+1,1})\\
=&[(\mathbf{x}_0+\mathbf{b}_{i,1})+UAT^R_i+\mathbf{b}_{i+1,2}]+\mathbf{W}_{i+1,2}'\sigma\{\mathbf{W}_{i+1,1}'[(\mathbf{x}_0+\mathbf{b}_{i,1})+UAT^R_i]+\mathbf{b}_{i+1,1}\}\\
=&(\mathbf{x}_0+\underline{\mathbf{b}_{i,1}+\mathbf{b}_{i+1,2}})+UAT^R_i+\mathbf{W}_{i+1,2}'\sigma\{\mathbf{W}_{i+1,1}'\mathbf{x}_0+\underline{(\mathbf{W}_{i+1,1}'\mathbf{b}_{i,1}+\mathbf{b}_{i+1,1})+\mathbf{W}_{i+1,1}'UAT^R_i}\}\\
\end{aligned}
\label{eq:UAT-res-cnn-2}
\end{equation}
\end{figure*}

\begin{figure*}[htbp!]
\centering
\includegraphics[width=0.9\textwidth]{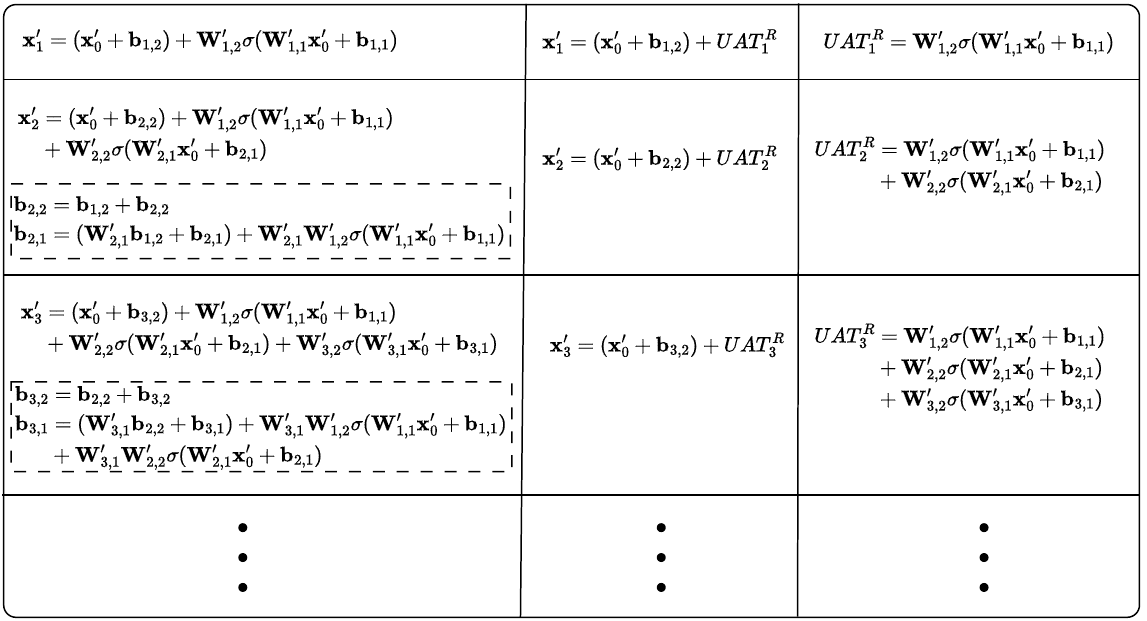}
\caption{Some examples of the UAT format of multi-layer residual-based convolution. The changes of parameters are represented within the dashed boxes. The parameters on the right of the equations indicate the original values, while those on the left represent the transformed values. There is no specific order of calculation for the parameters within each dashed box, but there is a top-to-bottom calculation order between different dashed boxes.
}
\label{fig:Conv_UAT}
\end{figure*}

\subsection{The DUAT Format of Transformer-based ViTs}
\label{section:The UAT Format of Transformer-based ViTs}

Similarly, to prove that Transformer-based ViTs are also DUAT functions, we start by deriving their general form based on Figure 9 in the main text. Transformers involve two key operations: Multi-Head Attention (MHA) and the Feed-Forward Network (FFN). In matrix-vector form, these operations correspond to Eqs. \ref{eq:MHA} and \ref{eq:FFN}.

\begin{equation}
MHA(\mathbf{x}_{i}) \mapsto \mathbf{W}'_{i,1}\mathbf{x}'_{i}
\label{eq:MHA}
\end{equation}

\begin{equation}
FFN(\mathbf{x}_{i}) \mapsto \mathbf{W}'_{i,3}\sigma (\mathbf{W}'_{i,2}\mathbf{x}'_{i}+\mathbf{b}'_{i,2})+\mathbf{b}'_{i,3}\\
\label{eq:FFN}
\end{equation}

\begin{figure*}[htbp!]
\centering
\begin{equation}
\centering
\begin{aligned}
\mathbf{x}'_{i+1} & =\mathbf{W}'_{i+1,1}\mathbf{x}'_{i}+\mathbf{W}'_{i+1,3}\sigma [\mathbf{W}'_{i+1,2}(\mathbf{W}'_{i+1,1}\mathbf{x}'_{i})+\mathbf{b}'_{i+1,2}]+\mathbf{b}'_{i+1,3}\\
&=\mathbf{W}'_{i+1,1}\mathbf{x}'_{i}+\mathbf{W}'_{i+1,3}\sigma (\underline{ \mathbf{W}'_{i+1,2}\mathbf{W}'_{i+1,1}}\mathbf{x}'_{i}+\mathbf{b}'_{i+1,2})+\mathbf{b}'_{i+1,3}\\
\end{aligned}
\label{eq:TF-general-term}
\end{equation}
\end{figure*}

Thus, a general term in a Transformer-based network can be expressed as Eq. \ref{eq:TF-general-term}. Letting $\mathbf{W}'_{i+1,2} = \mathbf{W}'_{i+1,2} \mathbf{W}'_{i+1,1}$, we can rewrite the general term as follows:

\begin{equation}
\mathbf{x}'_{i+1} = (\mathbf{W}'_{i+1,1} \mathbf{x}'_{i} + \mathbf{b}'_{i+1,3}) + \mathbf{W}'_{i,3} \sigma (\mathbf{W}'_{i+1,2} \mathbf{x}'_{i} + \mathbf{b}'_{i+1,2})
\label{eq:TF-general-term-simplify}
\end{equation}

Clearly, Eq. \ref{eq:TF-general-term-simplify} matches the mathematical form presented in Section \ref{section:From UAT to DUAT}: $\mathbf{x}'_{i+1} = (\mathbf{W}'_{i+1,1} \mathbf{x}'_{i} + \mathbf{b}'_{i+1,3}) + \mathbf{W}'_{i+1,3} \sigma (\mathbf{W}'_{i+1,2} \mathbf{x}'_{i} + \mathbf{b}'_{i+1,2})$. Therefore, a multilayer Transformer-based ViT is indeed DUAT function.

To clearly show the DUAT format of a multilayer ViT, we use the approach outlined in Section \ref{section:From UAT to DUAT}. Assuming that the overall mathematical form of $\mathbf{x}'_i$ aligns with the UAT structure, we decompose $\mathbf{x}'_i$ into a main term and a residual term, expressed as $\mathbf{x}'_i = (\mathbf{W}_{i,1} \mathbf{x}_0 + \mathbf{b}_{i,3}) + UAT^R_i$, where $UAT_{i}^R = \sum_{j=1}^{i} \mathbf{W}'_{j,3} \sigma (\mathbf{W}'_{j,2} \mathbf{x}'_{0} + \mathbf{b}'_{j,2})$ and $i=1,2,\ldots$. Substituting this into $\mathbf{x}'_{i+1} = (\mathbf{W}'_{i+1,1} \mathbf{x}'_{i} + \mathbf{b}'_{i+1,3}) + \mathbf{W}'_{i,3} \sigma (\mathbf{W}'_{i+1,2} \mathbf{x}'_{i} + \mathbf{b}'_{i+1,2})$, we get Eq. \ref{eq:UAT-TF}.

\begin{figure*}[htbp!]
\centering
\begin{equation}
\begin{aligned}
\mathbf{x}'_{i+1}&= (\mathbf{W}'_{i+1,1}\mathbf{x}'_{i}+\mathbf{b}'_{i+1,3})+\mathbf{W}'_{i,3}\sigma (\mathbf{W}'_{i+1,2}\mathbf{x}'_{i}+\mathbf{b}'_{i+1,2})\\
&= \{\mathbf{W}'_{i+1,1}[(\mathbf{W}_{i,1}\mathbf{x}_0+\mathbf{b}_{i,1})+UAT^R_i]+\mathbf{b}'_{i+1,3}\}\\
&+\mathbf{W}'_{i,3}\sigma \{\mathbf{W}'_{i+1,2}[(\mathbf{W}_{i,1}\mathbf{x}_0+\mathbf{b}_{i,1})+UAT^R_i]+\mathbf{b}'_{i+1,2}\}\\
&= [\underline{ \mathbf{W}'_{i+1,1}\mathbf{W}_{i,1}}\mathbf{x}_0+\underline{(\mathbf{W}'_{i+1,1}\mathbf{b}_{i,1}+\mathbf{b}'_{i+1,3})}]+\mathbf{W}'_{i+1,1}UAT^R_i]\\
&+\mathbf{W}'_{i,3}\sigma \{\underline{ \mathbf{W}'_{i+1,2}\mathbf{W}_{i,1}}\mathbf{x}_0+\underline{ (\mathbf{W}'_{i+1,2}\mathbf{b}_{i,1}+\mathbf{b}'_{i+1,2})+\mathbf{W}'_{i+1,2}UAT^R_i}\}\\
\end{aligned}
\label{eq:UAT-TF}
\end{equation}
\end{figure*}

Let $\mathbf{W}'_{i+1,1} = \mathbf{W}'_{i+1,1} \mathbf{W}_{i,1}$, $\mathbf{b}'_{i+1,3} = (\mathbf{W}'_{i+1,1} \mathbf{b}_{i,1} + \mathbf{b}'_{i+1,3})$, $UAT^R_i = \mathbf{W}'_{i+1,1} UAT^R_i$, $\mathbf{W}'_{i+1,2} = \mathbf{W}'_{i+1,2} \mathbf{W}_{i,1}$, and $\mathbf{b}'_{i+1,2} = (\mathbf{W}'_{i+1,2} \mathbf{b}_{i,1} + \mathbf{b}'_{i+1,2}) + \mathbf{W}'_{i+1,2} UAT^R_i$. Substituting these, we obtain $\mathbf{x}'_{i+1} = (\mathbf{W}'_{i+1,1} \mathbf{x}_0 + \mathbf{b}'_{i+1,3}) + UAT^R_i + \mathbf{W}'_{i,3} \sigma (\mathbf{W}'_{i+1,2} \mathbf{x}_0 + \mathbf{b}'_{i+1,2})$. Since $UAT^R_i$ overall aligns with the UAT, the overall mathematical form of the $i+1$ layers' ViT also conforms to the UAT.

So the $i+1$ layers' ViT can be expressed as:
\begin{equation}
\begin{aligned}
\mathbf{x}_{i+1}=(\mathbf{W}_{i+1,1}'\mathbf{x}_0+\mathbf{b}_{i+1,3})+\sum_{j=1}^{i+1}\mathbf{W}'_{j,3}\sigma (\mathbf{W}'_{j,2}\mathbf{x}'_{0}+\mathbf{b}'_{j,2})
\end{aligned}
\label{eq:TF-app}
\end{equation}

When $ i = 0 $, the parameters are defined as follows: $ \mathbf{W}'_{1,1} = \mathbf{W}'_{1,1} $, $ \mathbf{b}'_{1,3} = \mathbf{b}'_{1,3} $, $ \mathbf{W}'_{1,3} = \mathbf{W}'_{1,3} $, $ \mathbf{W}'_{1,2} = \mathbf{W}'_{1,2} \mathbf{W}'_{1,1} $, and $ \mathbf{b}'_{1,2} = \mathbf{b}'_{1,2} $. We define the parameters for cases where $ i \geq 1 $. For each $ j = 1, 2, \ldots, i $, the updates are as follows: $ \mathbf{W}'_{j+1,1} = \mathbf{W}'_{j+1,1} \mathbf{W}_{j,1} $, $ \mathbf{b}'_{j+1,3} = \mathbf{W}'_{j+1,1} \mathbf{b}_{j,3} + \mathbf{b}'_{j+1,3} $, $ \mathbf{W}'_{j+1,2} = \mathbf{W}'_{j+1,2} \mathbf{W}_{j,1}$, and $ \mathbf{W}'_{j,3} = \mathbf{W}'_{j+1,1} \mathbf{W}'_{j,3}$. 

Additionally, for $ j = 1, 2, \ldots, i + 1 $, the bias terms $ \mathbf{b}'_{j,2} $ are updated according to:
\begin{equation}
\begin{aligned}
\mathbf{b}'_{j,2} =& (\mathbf{W}'_{j,2} \mathbf{b}'_{j-1,3} + \mathbf{b}'_{j,2}) \\
+& \mathbf{W}'_{j,2} \sum_{k=1}^{j-1} \mathbf{W}'_{k,3} \sigma (\mathbf{W}'_{k,2} \mathbf{x}'_{0} + \mathbf{b}'_{k,2}).
\end{aligned}
\label{eq:b_j2-app}
\end{equation}

It means that the whole computing process is serial, if we want to compute $\mathbf{W}'_{3,1}$, we must calculate $\mathbf{W}'_{2,1}$, because $\mathbf{W}'_{3,1}=\mathbf{W}'_{3,1}\mathbf{W}'_{2,1}$. Thus, we have established that a multilayer ViT is also a DUAT function. The mathematical representation of the ViT aligns with the DUAT framework, with $\mathbf{b}'_{j,2}$ closely approximated by DUAT. The most notable difference between ViTs and residual-based CNNs is the dynamic input-dependent nature of the parameters in the MHA mechanism. Consequently, in the DUAT formulation associated with ViT, the parameters $\mathbf{W}'_{j,1}$ and $\mathbf{W}'_{j,2}$ for $i+1 \geq j \geq 1$, and $\mathbf{W}'_{j,3}$ for $i+1 > j > 1$ in layer $i$, all adapt dynamically based on the input.

\begin{figure*}[htbp!]
\centering
\includegraphics[width=0.9\textwidth]{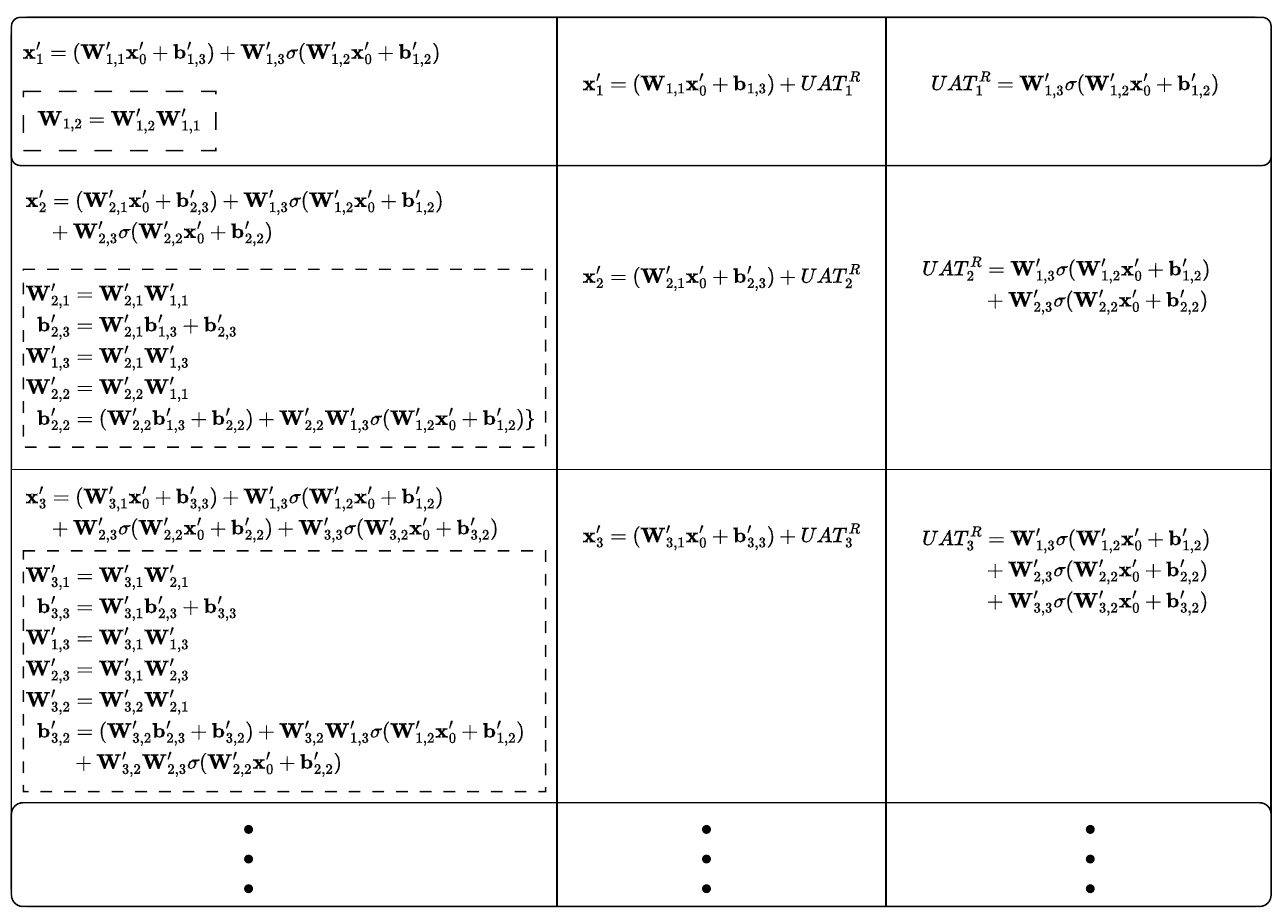}
\caption{Some examples of the UAT format of multi-layer Transformer. The changes of parameters are represented within the dashed boxes. The parameters on the right of the equations indicate the original values, while those on the left represent the transformed values. There is no specific order of calculation for the parameters within each dashed box, but there is a top-to-bottom calculation order between different dashed boxes.
}
\label{fig:TF_UAT}
\end{figure*}

\end{document}